\documentclass[11pt,a4paper]{article}
\usepackage[hyperref]{emnlp2018}
\usepackage{times}
\usepackage{latexsym}
\usepackage{caption}
\usepackage{url}
\usepackage[titletoc,title]{appendix}
\usepackage{soul}
\usepackage{tabularx}
\usepackage{graphicx}
\usepackage{caption}
\DeclareCaptionFont{small}{\small}
\usepackage{amsthm}
\theoremstyle{definition}
\newtheorem{definition}{Definition}[section]
\theoremstyle{remark}

\usepackage{amssymb}
\usepackage{url}
\usepackage{makecell}

\usepackage[T1,T5]{fontenc}
\usepackage{kotex}
\usepackage{array,multirow}
\usepackage{booktabs}
\usepackage{CJKutf8}  
\usepackage{makecell}

\aclfinalcopy 

\title{
  Massively Parallel Cross-Lingual Learning \\
  in Low-Resource Target Language Translation
}

\author{
  Zhong Zhou\\
  {\small Carnegie Mellon University}\\
  {\small \tt zhongzhou@cmu.edu} 
  \\\And 
  Matthias Sperber\\
  {\small Karlsruhe Institute of Technology}\\
  {\small \tt matthias.sperber@kit.edu}
  \\\And 
  Alex Waibel\\
  \small Carnegie Mellon University\\
  \small Karlsruhe Institute of Technology\\
  {\small \tt alex@waibel.com }
}

\date{}

\begin{document}

\maketitle

\begin{abstract}
  We work on translation from rich-resource languages to
  low-resource languages. The main challenges we 
  identify are
  the lack of low-resource language data, 
  effective methods for cross-lingual transfer,
  and the variable-binding problem that is common in
  neural systems. We build a translation system
  that addresses these challenges using eight 
  European language families as our test ground. 
  Firstly, we add the source and the target family 
  labels and study
  intra-family and inter-family influences 
  for effective cross-lingual transfer.
  We achieve an improvement of +9.9 in BLEU score
  for English-Swedish translation using eight families 
  compared to the single-family
  multi-source multi-target baseline. Moreover,
  we find that training on two neighboring 
  families closest to
  the low-resource language is often enough.
  Secondly, we construct an ablation study 
  and find that reasonably good results can be achieved 
  even with considerably less target data. 
  Thirdly, we address the variable-binding
  problem by building an order-preserving 
  named entity translation model. We obtain 60.6\% accuracy in 
  qualitative evaluation where
  our translations are akin to 
  human translations in a preliminary study. 
\end{abstract}

\section{Introduction} \label{introduction}
We work on translation 
from a rich-resource language 
to a low-resource language. There is usually 
little low-resource
language data, much less parallel data available
\cite{duong2016attentional, anastasopoulos2017spoken};
Despite of the challenges of little data
and few human experts, 
it has many 
useful applications. Applications include translating 
water, sanitation and
hygiene (WASH) guidelines to protect 
Indian tribal children against 
waterborne diseases, introducing 
earthquake preparedness techniques to Indonesian 
tribal groups living near volcanoes 
and delivering information to the disabled or the
elderly in low-resource language communities 
\cite{reddy2017water, barrett2005support, anastasiou2010translating, perry2017treasure}.
These are useful examples of 
translating a closed text known in advance 
to the low-resource language. 

There are three main challenges. 
Firstly, most of previous works research on
individual languages instead of collective 
families. Cross-lingual
impacts and similarities are very helpful 
when there is little data
in low-resource language 
\cite{shoemark2016towards, sapir1921languages, odlin1989language, cenoz2001effect, toral2018level, de1997pursuit, hermans2003cross, specia2016shared}.
Secondly, 
most of the multilingual Neural Machine 
Translation (NMT) works assume
the same amount of training data for all 
languages. In the low-resource case,
it is important to exploit low or partial data
in low-resource language to produce high
quality translation. 
The third issue is the variable-binding
problem that is common in neural systems, 
where ``John calls Mary''
is treated the same way as ``Mary calls John'' 
\cite{fodor1988connectionism, graves2014neural}.
It is more challenging when both ``Mary'' 
and ``John'' are rare words. Solving 
the binding problem is crucial because the 
mistakes in 
named entities change the meaning of 
the translation. It is especially challenging
in the low-resource case because
many words are rare words.

Our contribution in addressing these issues 
is three-fold, extending from multi-source
multi-target attentional NMT.
Firstly, to examine intra-family and
inter-family influences, we add source and target 
language family labels in training.
Training on multiple families improves
BLEU score significantly; moreover, we find training on two 
neighboring families closest to the 
low-resource language gives reasonably
good BLEU scores, and we define neighboring families closely in Section
\ref{proposedextension}. Secondly, we conduct an 
ablation study to explore
how generalization changes with 
different amounts of data and
find that we only need a small amount of
low-resource language data to produce 
reasonably good BLEU scores. We use full data	
except for the ablation	study.
Finally, to address the variable-binding 
problem, we
build a parallel lexicon table across 
twenty-three European
languages and devise a novel 
method of order-preserving
named entity translation method. Our method works
in translation of any text with a fixed set of
named entities known in advance. 
Our goal is to minimize manual labor, but
not to fully automate to ensure
the correct translation of named entities and their
ordering. 

In this paper, we begin with introduction and 
related work in Section 
\ref{introduction} and \ref{relatedwork}. 
We introduce our methods in addressing 
three issues that are important for
translation into low-resource
language in Section \ref{proposedextension},
as proposed extensions to our baseline in
Section \ref{baseline}.
Finally, we present our results 
in Section \ref{experiments} and
conclude in Section \ref{conclusion}.

\section{Related Work} \label{relatedwork}
\subsection{Multilingual Attentional NMT}
Attentional NMT is trained directly in an end-to-end
system and has flourished recently \cite{wu2016google, sennrich2016neural, ling2015character}.
Machine polyglotism, training machines to be proficient
in many languages, is a new paradigm of multilingual NMT 
\cite{johnson2017google, ha2016toward, firat2016multi, zoph2016multi, dong2015multi, gillick2016multilingual, al2013polyglot, tsvetkov2016polyglot}.
Many multilingual NMT systems involve multiple encoders
and decoders, and it is hard to combine 
attention for quadratic language pairs bypassing 
quadratic attention mechanisms 
\cite{firat2016multi}.  
In multi-source scenarios, multiple
encoders share a combined attention
mechanism \cite{zoph2016multi}. In
multi-target scenarios, every decoder
handles its own attention
with parameter sharing \cite{dong2015multi}. 
Attention combination schemes include simple
combination and hierarchical combination
\cite{libovicky2017attention}.

The state-of-the-art of multilingual NMT is adding
source and target language labels in training
a universal model with a single attention scheme, and
Byte-Pair Encoding (BPE) is used at preprocessing stage \cite{ha2016toward}.
This method is elegant in its simplicity and its advancement in
low-resource language translation as well as zero-shot translation using
pivot-based translation scheme \cite{johnson2017google}. 
However, these works have training data that increases 
quadratically with the number of languages 
\cite{dong2015multi, gillick2016multilingual},  
rendering training on massively parallel corpora difficult. 

\subsection{Sub-word Level NMT}
Many NMT systems lack robustness with
out-of-vocabulary words (\textit{OOV}s) \cite{wu2016google}.
Most \textit{OOV}s are treated as unknowns (\textit{\$UNK}s) uniformly, 
even though they are semantically important and different  
\cite{ling2015character, sennrich2016neural}.
To tackle the \textit{OOV} problem,
researchers work on byte-level
\cite{gillick2016multilingual}
and character-level models
\cite{ling2015character, chung2016character}.
Many character-level models do not work as well
as word-level models, and do not produce optimal alignments
\cite{tiedemann2012character}. 
As a result, many researchers shift to sub-word level modeling between
character-level and word-level. One prominent direction is BPE which iteratively learns 
subword units and balances sequence length and expressiveness 
with robustness \cite{sennrich2016neural}.

\subsection{Lexiconized NMT} 
Much research is done in translating
lexicons and named entities in NMT 
\cite{nguyen2017improving, wang2017sogou, arthur2016incorporating}. 
Some researchers create a separate
character-level named entity model 
and mark all named entities as \textit{\$TERM}s
to train \cite{wang2017sogou}.
This method learns people's names well
but does not improve 
BLEU scores \cite{wang2017sogou}. It is
time-consuming and adds to the system complexity.
Other researchers attempt to build lexicon
translation seamlessly with attentional NMT
by using an affine transformation 
of attentional weights
\cite{nguyen2017improving, arthur2016incorporating}. 
Some also attempt to embed cross-lingual lexicons
into the same vector space for transfer
of information \cite{duong2017multilingual}. 
\begin{table}[t]
  \small
  \begin{tabular}{ | p{1.15cm} | p{5.8cm} | }
    \hline
    Families & Languages \\ \hline \hline
    Germanic & German (de) Danish (dn) Dutch (dt) Norwegian (no) Swedish (sw) English (en) \\\hline
    Slavic & Croatian (cr) Czech (cz) Polish (po) Russian (ru) Ukrainian (ur) Bulgarian (bg) \\\hline
    Romance & Spanish (es) French (fr) Italian (it) Portuguese (po) Romanian (ro) \\\hline
    Albanian & Albanian (ab) \\\hline
    Hellenic & Greek (gk) \\\hline
    Italic & Latin (ln) [descendants: Romance languages] \\\hline
    Uralic & Finnish (fn) Hungarian (hg) \\\hline
    Celtic & Welsh (ws) \\ \hline
  \end{tabular}
  \caption{Language families. Language codes are in brackets.}
  \label{table:family}
\end{table}

\section{Translation System}

\subsection{Baseline Translation System} \label{baseline}
Our baseline is multi-source multi-target attentional NMT within
one language family through adding
source and target language labels with a single
unified attentional scheme, with BPE used at the preprocessing stage.
The source and target vocabulary are not shared. 

\subsection{Proposed Extensions}\label{proposedextension}
We present our methods in solving three
issues relevant to translation into low-resource language as
our proposed extensions. 

\subsubsection{Language Families and Cross-lingual Learning} 
Cross-lingual and cross-cultural influences and similarities
are important in linguistics 
\cite{shoemark2016towards, levin1998interlingua, sapir1921languages, odlin1989language, cenoz2001effect, toral2018level, de1997pursuit, hermans2003cross, specia2016shared}. 
The English word, ``Beleaguer'' originates
from the Dutch word ``belegeren''; ``fidget''
originates from the Nordic word ``fikja''.
English and Dutch belong to the same family
and their proximity has effect on
each other
\cite{harding1988classification, ross2006language}.
Furthermore, languages that do not belong
to the same family affect each other too 
\cite{sapir1921languages, ammon2001dominance, toral2018level}. 
``Somatic'' stems from the Greek
word ``soma''; ``\begin{CJK}{UTF8}{min}広告\end{CJK}'' (Japanese), ``광고''(Korean), ``Qu{\h{a}}ng c{\'a}o''(Vietnamese)
are closely related to the Traditional Chinese word ``\begin{CJK*}{UTF8}{bsmi}{\CJKfamily{bkai}廣告}\end{CJK*}''. Indeed, 
many cross-lingual similarities are present. 

In this paper, we use the language phylogenetic tree
as the measure of closeness of languages and language
families \cite{petroni2008language}. The distance measure
of language families is the collective of all of the
component languages. Language families that are next to each other
in the language phylogenetic tree are treated as neighboring
families in our paper, like Germanic family and Romance family. 
In our discussion in this paper,
we will often refer to closely related families
in the language phylogenetic tree as neighboring
families. 

We prepend the source and target family labels,
in addition to the source and target language labels
to the source sentence to improve convergence
rate and increase translation performance. For example, all French-to-English
translation pairs are prepended with four labels, the source and target family labels
and the source and target languages labels, i.e.,
\texttt{\_\_opt\_family\_src\_romance \_\_opt\_family\_tgt\_germanic \_\_opt\_src\_fr \_\_opt\_tgt\_en}. 
In 
Section \ref{experiments}, we examine intra-family
and inter-family effects more closely. 

\subsubsection{Ablation Study on Target Training data} 
To achieve high information transfer from 
rich-resource language to low-resource target language, 
we would like to find out how much target training 
data is needed to produce reasonably good performance. 
We vary the amount of low-resource training data to 
examine how to achieve reasonably good BLEU score using limited low-resource data. 
In the era of Statistical Machine Translation (SMT), researchers have 
worked on data sampling and sorting measures 
\cite{eck2005low, axelrod2011domain}.

To rigorously determine how much low-resource 
target language is needed for reasonably good results, 
we do a range of control experiments by drawing samples from the
low-resource language data
randomly with replacement and duplicate
them if necessary to ensure all experiments carry
the same number of training sentences.
We keep the amount
of training data in rich-resource languages
the same, and vary
the amount of training data in low-resource language
to conduct rigorous control experiments.
Our data selection process is different
from prior research in that only the
low-resource training data is reduced,
simulating the real world scenario of having
little data in low-resource language. By 
comparing results from control
experiments, we determine how much 
low-resource data is needed. 

\subsubsection{Order-preserving Lexiconized NMT}
The variable-binding problem is an inherent issue in connectionist architectures \cite{fodor1988connectionism, graves2014neural}. ``John calls Mary'' 
is not equivalent to ``Mary calls John'', but neural networks 
cannot distinguish the two easily 
\cite{fodor1988connectionism, graves2014neural}.
The failure of traditional NMT to distinguish
the subject and the object of a sentence is detrimental.
For example, in the narration ``John told his son Ryan to
help David, the brother of Mary'',
it is a serious mistake if we reverse John and
Ryan's father-son relationships or
confuse Ryan's and David's relationships with Mary.  

In our research on translation, we focus mainly on text with
a fixed set of named entities known in advance. We assume that experts help to translate
a given list of named entities into low-resource language first before
attempting to translate any text. Under this assumption, we
propose an order-preserving named entity translation mechanism.
Our solution is to first create a parallel lexicon table for 
all twenty-three European languages using a seed English lexicon table
and fast-aligning it with the rest \cite{dyer2013simple}. 
Instead 
of using \textit{\$UNK}s to replace the named entities, we
use \textit{\$NE}s to distinguish them from the other unknowns.
We also sequentially tag named entities in a sentence
as \textit{\$NE1}, \textit{\$NE2}, \ldots, to preserve their ordering. 
For every sentence pair in the multilingual training,
we build a target named entity decoding dictionary by using all target
lexicons from our lexicon table that matches with those appeared in the
source sentence. During the evaluation stage, we 
replace all the numbered \textit{\$NE}s
using the target named entity decoding dictionary
to present our final translation.
This method improves translation accuracy greatly and preserves the order.

As a result of our contribution, the experts only need
to translate a few lexicons and a small amount of low-resource text before passing
the task to our system to obtain good results. Post-editing and minor changes
may be required to achieve 100\% accuracy before the releasing the
translation to the low-resource language communities. 

\section{Experiments and Results}\label{experiments}
\begin{table}[t]
    \small
    \centering
    \begin{tabularx}{\columnwidth}{|X|X|X|X|X|X|X|} \hline
      lan & de & dn & dt & en & no & sw\\ \hline \hline
      de & N.A. & 37.5 & 43.4 & 45.1 & 41.1 & 35.8\\ \hline
      dn & 39.0 & N.A. & 37.1 & 41.1 & 42.6 & 37.4\\ \hline
      dt & 43.5 & 36.3 & N.A. & 45.1 & 39.0 & 34.3\\ \hline
      en & 40.4 & 34.5 & 41.1 & N.A. & 37.1 & 34.0\\ \hline
      no & 40.5 & 42.7 & 40.4 & 42.8 & N.A. & 40.6\\ \hline
      sw & 39.4 & 38.9 & 37.5 & 40.4 & 43.0 & N.A.\\ \hline
    \end{tabularx}
    \caption{(Baseline model) Germanic family multi-source multi-target translation.
        Each row represents source, each column represents target.
        Language codes follow Table \ref{table:family}.
    }
    \label{table:germanic}
\end{table}
We choose the Bible corpus as a test ground for
our proposed extensions because the Bible is the most
translated text that exists and is freely
accessible. Though it has limitations, it does not have
copyright issues like most of literary works
that are translated into many languages do. 
There are many research works done using the Bible
\cite{naaijer1993parallel, mayer2014creating, scannell2006machine, dufter2018universal, resnik1999bible, chan2001encyclopaedia, banchs2011semantic, christodouloupoulos2015massively, beale2005document}.
Unlike many past research works where
only New Testament is used \cite{dufter2018universal},
we use both Old Testament and New Testament in our Bible corpus.   
We align all Bible verses across different languages.

We train our proposed model on twenty-three European languages 
across eight families on a parallel Bible corpus. 
For our purpose, we treat Swedish as our hypothetical 
low-resource target language, English as our
rich-resource language in the single-source single-target case and
all other Germanic languages as our rich-resource languages
in the multi-source multi-target case. 

Firstly, we present our data and training parameters. Secondly, we add
family tags in different configurations of language families showing
intra-family and inter-family effects.
Thirdly, we conduct an ablation study and plot the generalization curves
by varying the amount of training data in Swedish, and
we show that training on one fifth of the data give reasonably good BLEU scores.
Lastly, we devise an order-preserving lexicon translation method by
building a parallel lexicon table across twenty-three
European languages and tagging named entities in order.

\subsection{Data and Training Parameters} 
We clean and align the Bible in twenty-three European languages
in Table \ref{table:family}. We randomly sample the
training, validation and test sets according to the 0.75, 0.15, 0.10 ratio.
Our training set contains 23K verses, but is massively parallel. In our 
control experiments, we also use the experiment training on the WMT'14 French-English dataset
together with French and English Bibles to compare with 
our results. Note that our WMT baseline contains French and English Bibles
in addition to the WMT'14 data, and is used to contrast our results with
the effect of increasing data. 

In all our experiments, we use a minibatch size of 64, dropout rate of 0.3,
4 RNN layers of size 1000, a word vector size of 600, learning rate of 0.8
across all LSTM-based multilingual experiments. For single-source single-target translation, we use 2 RNN layers of
size 500, a word vector size of 500, and learning rate of 1.0. All learning
rates are decaying at the rate of 0.7 if the validation score is not improving
or it is past epoch 9. We use SGD as our learning algorithm.
We build our code based on OpenNMT \cite{klein2017opennmt}. 
For the ablation study, we train on BLEU scores directly 
until the \textit{Generalization Loss} (\textit{GL}) exceeds a threshold of $\alpha = 0.1$
\cite{prechelt1998early}. \textit{GL} at epoch $t$ is defined as
$GL(t) = 100 ( 1- \frac{E_{val}^t}{E_{opt}^t} )$,
modified by us to suit our objective using BLEU scores \cite{prechelt1998early}.
$E_{val}^t$ is the validation score at
epoch $t$ and $E_{opt}^t$ is the optimal score up to epoch $t$.
We evaluate our models using both BLEU scores \cite{papineni2002bleu} and qualitative evaluation.

\subsection{Family labels and Intra-family \& Inter-family Effects}
\begin{table}[t]
    \small
    \centering
    \begin{tabularx}{\columnwidth}{|X|X|X|X|X|X|X|} \hline
      expt & S & G & GS & GR & 3F & 8F  \\ \hline \hline
      de2sw & 4.0 & 35.8 & 42.0 & 42.2 & 42.5 & 42.8 \\ \hline
      dn2sw & 16.9 & 37.4 & 43.4 & 41.8 & 42.7 & 41.7 \\ \hline
      dt2sw & 4.8 & 34.3 & 41.4 & 41.6 & 42.8 & 42.5 \\ \hline
      en2sw & 6.9 & 34.0 & 40.3 & 40.2 & 41.8 & 42.1 \\ \hline
      no2sw & 16.8 & 40.6 & 43.6 & 44.0 & 44.5 & 43.1 \\ \hline
    \end{tabularx}
    \caption{Inter-family and intra-family effects on BLEU scores with respect to increasing addition of language families. \\
      S: single-source single-target NMT.
      \\G: training on Germanic family.
      \\GS: training on Germanic, Slavic family.
      \\GR: training on Germanic, Romance family.
      \\3F: training on Germanic, Slavic, Romance family.
      \\8F: training on all 8 European families together.
    }
    \label{table:summary}
\end{table}
\begin{figure}[t]
  \centering
  \includegraphics[width=3.1in]{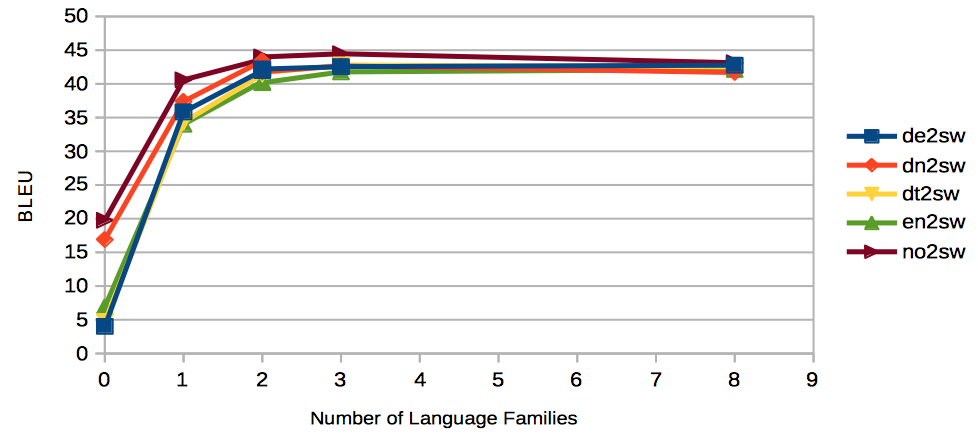}
  \caption{Intra-family and inter-family effects on BLEU scores with respect to increasing addition of language families. }
  \label{fig:multi_family1}
\end{figure}
\begin{table}[t]
    \small
    \centering
    \begin{tabularx}{\columnwidth}{|X|X|X|X|X|X|X|} \hline
      expt & S & G & GSl & GRl & 3Fl & 8Fl  \\ \hline \hline
      de2sw & 4.0 & 35.8 & 41.8 & 42.2 & 42.5 & 44.3 \\ \hline
      dn2sw & 16.9 & 37.4 & 43.0 & 41.5 & 42.5 & 42.8 \\ \hline
      dt2sw & 4.8 & 34.3 & 41.4 & 41.8 & 42.7 & 42.3 \\ \hline
      en2sw & 6.9 & 34.0 & 40.9 & 40.4 & 41.7 & 43.9 \\ \hline
      no2sw & 16.8 & 40.6 & 43.7 & 44.3 & 44.2 & 44.7 \\ \hline
    \end{tabularx}
    \caption{Effects of adding family labels on BLEU scores with respect to increasing addition of language families. \\
      S and G: same as in Table \ref{table:summary}.
      \\GSl: Germanic, Slavic family with family labels.
      \\GRl: Germanic, Romance family with family labels.
      \\3Fl: Germanic, Slavic, Romance family with family labels.
      \\8Fl: all 8 European families together with family labels
    }
    \label{table:summary_label}
\end{table}
\begin{figure}[t]
  \centering
  \includegraphics[width=3.1in]{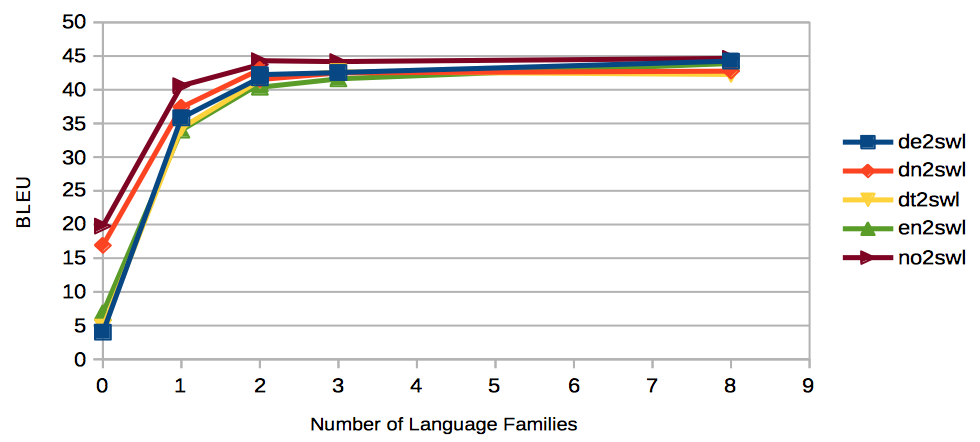}
  \caption{Effects of adding family labels on BLEU scores with respect to increasing addition of language families. }
  \label{fig:multi_family2}
\end{figure}
We first investigate intra-family
and inter-family influences and the
effects of adding family labels.
We use full training data
in this subsection. Adding family labels
not only improves convergence rate,
but also increases BLEU scores. 

\textbf{\ul{Languages have varying closeness to each other:}}
Single-source single-target translations of different languages in
Germanic family to Swedish show huge differences in BLEU scores as
shown in Table \ref{table:summary}. These differences are well
aligned with the multi-source multi-target results. Norwegian-Swedish
and Danish-Swedish translations have much higher BLEU scores
than the rest. This hints that Norwegian and Danish are closer to Swedish
than the rest in the neural representation. 

\textbf{\ul{Multi-source multi-target translation improves greatly from single-source
    single-target translation:}} English-Swedish
single-source single-target translation gives a low BLEU score
of 6.9 as shown in Table \ref{table:summary},
which is understandable as our dataset is very small.
BLEU score for English-Swedish translation
improves greatly to 34.0 in multi-source multi-target
NMT training on Germanic family as shown in
Table \ref{table:germanic}. In this paper, we treat Germanic multi-source
multi-target NMT as our baseline model.
Complete tables of
multi-source and multi-target experiments
are in the appendices. We present only
relevant columns important for cross-lingual learning and translation
into low-resource language here.

\textbf{\ul{Adding languages from other families into training improves
    translation quality within each family greatly:}} English-Swedish translation's BLEU score
improves significantly from 34.0 to 40.3 training on Germanic and Slavic families,
and 40.2 training on Germanic and Romance
families as shown in Table \ref{table:summary}.
After we add all three families in training,
BLEU score for English-Swedish translation increases further to
41.8 in Table \ref{table:summary}. Finally,
after we add all eight families,
BLEU score for English-Swedish translation
increases to 42.1 in Table \ref{table:summary}.
\begin{figure}[t]
  \centering
  \includegraphics[width=3.1in]{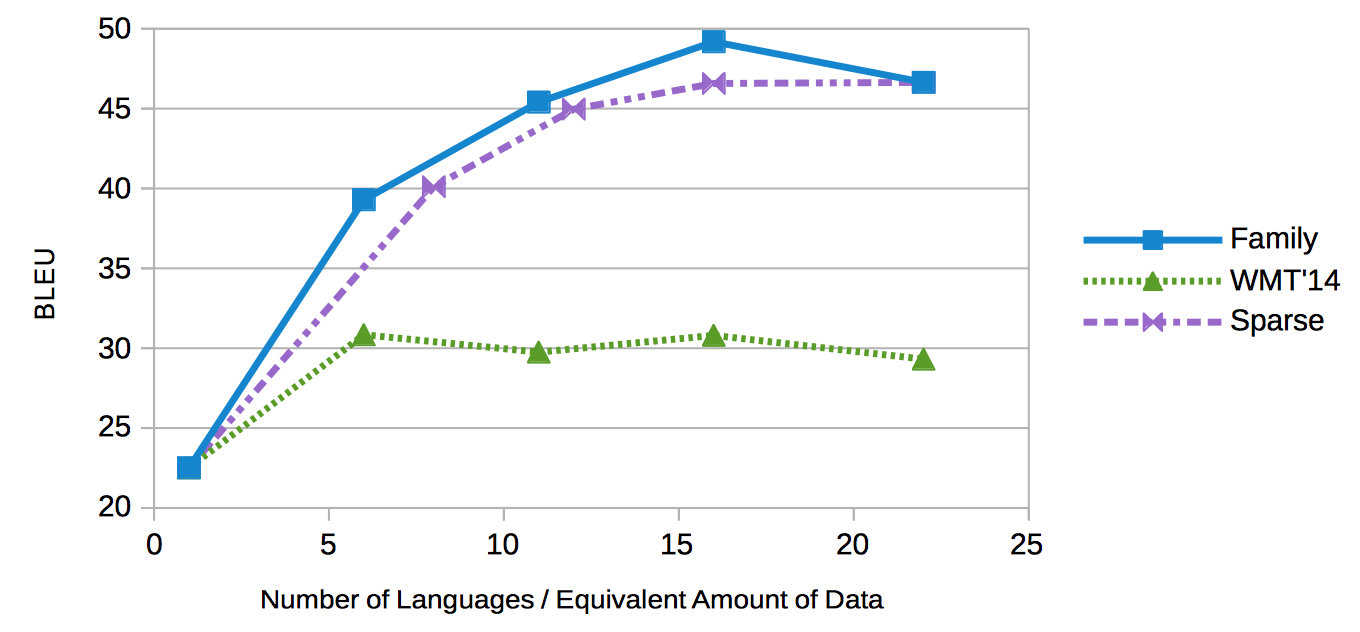}
  \caption{Comparison of different ways of increasing training Data in French-English translation. \\
      Family: Adding data from other languages based on the family unit
      \\WMT'14: Adding WMT'14 data as control experiment
      \\Sparse: Adding data from other languages that spans the eight European families
  }
  \label{fig:adding}
\end{figure}

\textbf{\ul{A Plateau is observed after adding more than one
    neighboring family:}} A plateau is observed when we plot
Table \ref{table:summary} in Figure \ref{fig:multi_family1}.
The increase in BLEU scores after adding two families is much milder
than that of the first addition of a neighboring family. This
hints that using unlimited number of languages to train may not
be necessary.

\textbf{\ul{Adding family labels not only improves convergence rate,
    but also increases BLEU scores:}} We observe in Table \ref{table:summary_label} that
BLEU scores for most language pairs improve with the addition of
family labels. Training on eight language families,
we achieve a BLEU score of 43.9 for English-Swedish
translation, +9.9 above the Germanic baseline. Indeed,
the more families we have, the more helpful it is to distinguish them. 

\textbf{\ul{Training on two neighboring families nearest to the
    low-resource language gives better result than training on languages
    that are further apart:}} Our observation of the
plateau hints that training on two neighboring
families nearest to the low-resource language
is good enough as shown in Table \ref{table:summary}.
Before jumping to conclusion, we compare results of adding languages
by family with that of adding languages by random samples that
span all eight families, defined as the following. 
\theoremstyle{definition}
\begin{definition}[Language Spanning]
  A set of languages spans a set of families when it contains
  at least one language from each family.
\end{definition}
In Figure \ref{fig:adding}, we conduct a few experiments 
on French-English translation using
different ways of adding training data. Let
\textit{family addition} describe the addition of training data 
through adding close-by language families 
based on the unit of family; let
\textit{sparse addition} describe 
the addition of training data through adding language sets that spans eight 
language families. In sparse addition, languages are further apart as each may 
represent a different family. We find that
family addition gives better generalization
than that of sparse addition.  
It strengthens our earlier results that 
training on two families closest
to our low-resource language is a 
reliable way to reach good generalization. 

\textbf{\ul{Generalization is not merely an effect of
    increasing amount of data:}} In Figure \ref{fig:adding}, we compare all methods of adding
languages against a WMT'14 curve by
using equivalent amount of
WMT'14 French-English data in each experiment.
The WMT'14 curve serve as our benchmark
of observing the effect of increasing data,
we observe that our addition of other
languages improve BLEU score much sharply
than the increase in the benchmark, showing that
our generalization is not merely an
effect of increasing data. We also observe
that though increase WMT'14 data initially increases
BLEU score, it reaches a plateau and adding more
WMT'14 data does not increase performance from
very early point.  
\begin{table*}[t]
    \small
    \centering
    \begin{tabularx}{\textwidth}{|X|X|X|X|X|X|X|X|X|X|X|} \hline
      Data & 0.1 & 0.2 & 0.3 & 0.4 & 0.5 & 0.6 & 0.7 & 0.8 & 0.9 & 1 \\ \hline
      \#w &53589 &107262 &161332 &214185 &268228 &322116 &375439 &429470 &483440 &538030 \\ \hline
      log\#w &4.73 &5.03 &5.21 &5.33&5.43 &5.51 &5.57 &5.63 &5.68 &5.73 \\ \hline \hline
      en2sw &25.2 &30.6 &32.9 &32.7 &34.2 &34.2 &33.8 &33.6 &34.3 &34.9 \\ \hline
      de2sw &26.5 &33.4 &34.8 &35.7 &36.7 &36.5 &37.1 &37.1 &36.4 &37.5 \\ \hline
      dn2sw &27.2 &34.8 &35.8 &37.1 &37.6 &37.1 &38.5 &38.0 &37.4 &38.4 \\ \hline
      dt2sw &26.1 &32.5 &34.2 &34.9 &36.0 &35.8 &36.0 &35.7 &35.8 &36.6 \\ \hline
      no2sw &27.7 &36.9 &37.9 &39.5 &39.4 &39.2 &41.3 &40.8 &39.2 &40.5 \\ \hline
    \end{tabularx}
    \caption{Ablation Study on Germanic Family. \#w is the word count of unique sentences in Swedish data. }
    \label{table:ablation}
\end{table*}
\begin{table}[t]
  \small
  \centering
  \begin{tabular}{ |p{0.66cm}|p{0.92cm}|p{0.96cm}|p{0.8cm}|p{0.92cm}|p{0.98cm}| }
      \hline
      en & de & cz & es & fn & sw \\ \hline \hline
      Joseph&Joseph&Jozef&Jos{\'e}&Joseph&Josef \\ \hline
      Peter&Petrus&Petr&Pedro&Pietari&Petrus \\ \hline
      Zion&Zion&Sion&Sion&Zionin&Sion \\ \hline
      John&Johannes&Jan&Juan&Johannes&Johannes \\ \hline
      Egypt&{\"A}gypten&Egyptsk{\'e}&Egipto&Egyptin&Egyptens \\ \hline
      Noah&Noah&No{\'e}&No{\'e}&Noa&Noa \\ \hline
  \end{tabular}
  \caption{A few examples from the parallel lexicon table.}
  \label{table:lexicon_examples}
\end{table}
\begin{table}[t]
    \small
    \centering
    \begin{tabularx}{\columnwidth}{|X|X|X|X|X|} \hline
      expt & G & OG & OG1 & OGM  \\ \hline \hline
      de2sw & 35.8 & 36.6 & 36.6 & 36.9 \\ \hline
      dn2sw & 37.4 & 37.0 & 37.2 & 36.9  \\ \hline
      dt2sw & 34.3 & 35.8 & 35.6 & 35.9 \\ \hline
      en2sw & 34.0 & 33.6 & 33.9 & 33.4  \\ \hline
      no2sw & 40.6 & 41.2 & 41.0 & 41.4  \\ \hline
    \end{tabularx}
    \caption{Summary of order-preserving lexicon translation. \\
      G: training on Germanic family without using order-preserving method. \\
      OG: order-preserving lexicon translation.
      \\OG1: OG translation using lexicons with frequency 1.
      \\OGM: OG translation using lexicons with manual selection.
    }
    \label{table:summary_lexicon}
\end{table}
\begin{figure}[t]
  \centering
  \includegraphics[width=3.1in]{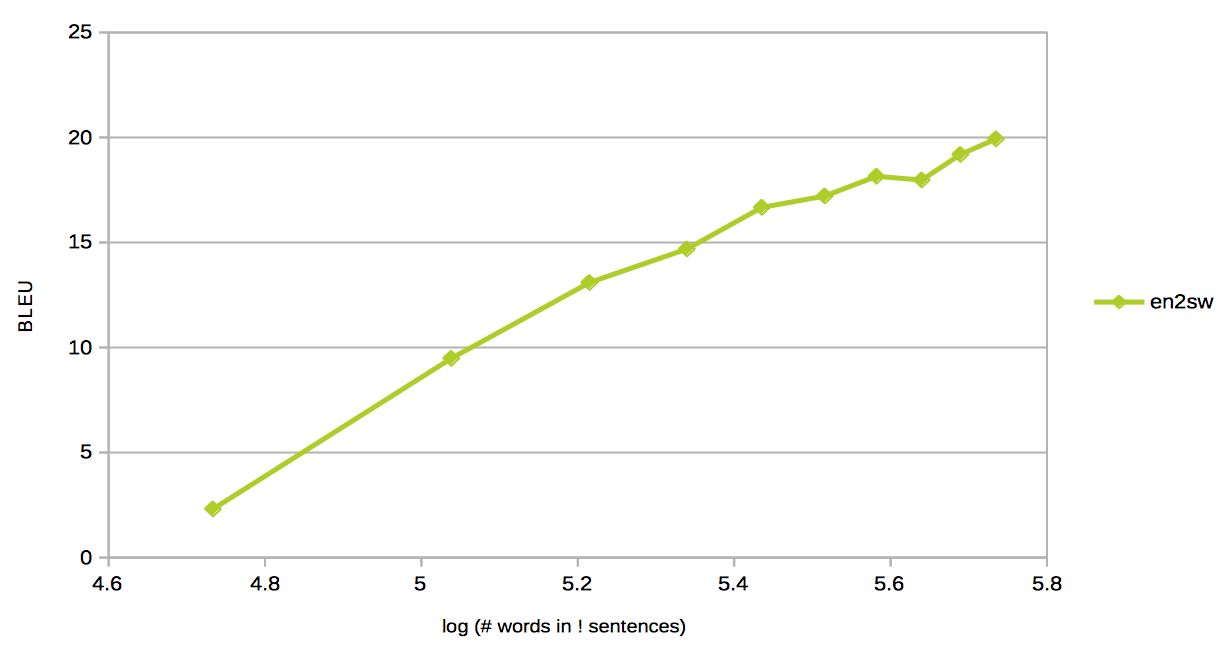}
  \caption{Single-source single-target English-Swedish BLEU plots against increasing amount of Swedish data.}
  \label{fig:curve_single}
\end{figure}
\begin{figure}[t]
  \centering
  \includegraphics[width=3.1in]{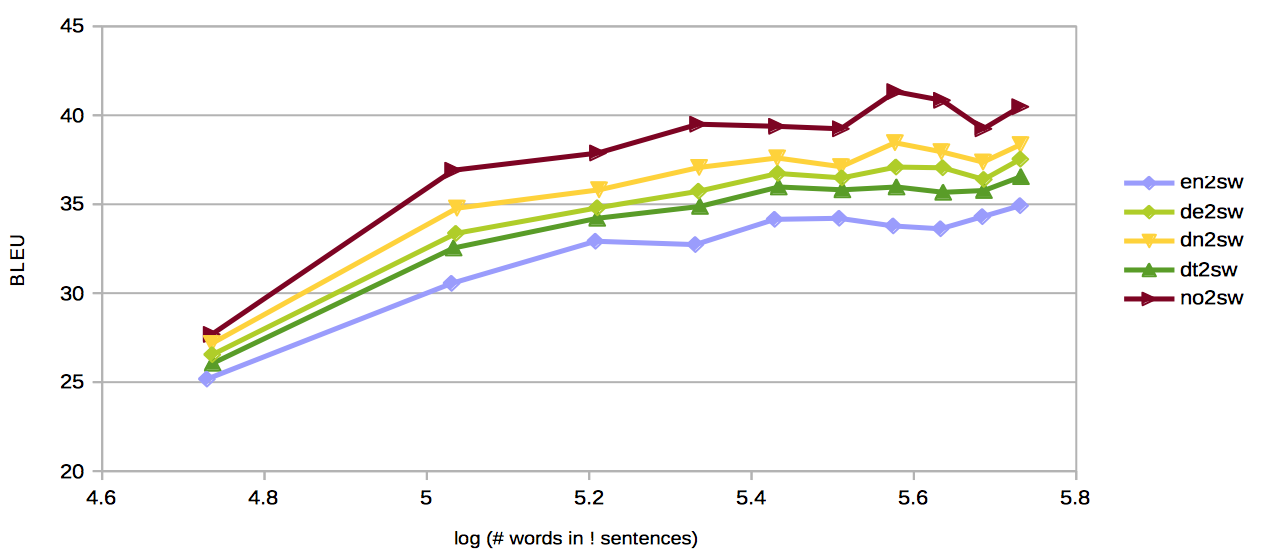}
  \caption{Multi-source multi-target
Germanic-family-trained BLEU plots
against increasing amount of Swedish
data.}
  \label{fig:curve_multi}
\end{figure}
\begin{table*}[]
  \small
  \centering
  \begin{tabular}{ |p{3.2cm}|p{2.8cm}|p{2.8cm}|p{2.8cm}|p{2.77cm}| } \hline
    Source Sentence
    & NMT Translation without Order Preservation (Before)
    & NMT Translation with Order Preservation (After)
    & Correct Target Translation
    & Frequency of Named Entities\\ \hline \hline
    And \emph{Noah} fathered three sons, \emph{Shem}, \emph{Ham}, and \emph{Japheth}.
      & Och \emph{Noa} f{\"o}dde tre s{\"o}ner, \emph{Sem}, \emph{Ham} och \emph{Jafet}.
      & Och \emph{Noa} f{\"o}dde tre s{\"o}ner, \emph{Sem} \emph{Ham} och \emph{Jafet}
      & Och \emph{Noa} f{\"o}dde tre s{\"o}ner: \emph{Sem}, \emph{Ham} och \emph{Jafet}.
      & \emph{Noah}: 58, \emph{Shem}: 18, \emph{Ham}: 17, \emph{Japheth}: 11
    \\ \hline
    And \emph{Saul} spoke to his son \emph{Jonathan}, and to all his servants, to kill \emph{David}.
      & Och \emph{Saul} sade till \emph{Jonatan}, hans son, och alla hans tj{\"a}nare, s{\aa} att de skulle d{\"o}da \emph{David}.
      & Och \emph{Saul} talade till sin son \emph{Jonatan} och alla hans tj{\"a}nare f{\"o}r att d{\"o}da \emph{David}
      & Och \emph{Saul} talade med sin son \emph{Jonatan} och med alla sina tj{\"a}nare om att d{\"o}da \emph{David}
      & \emph{Saul}: 424, \emph{Jonathan}: 121, \emph{David}: 1134
    \\ \hline
    And they killed \emph{Parshandatha}, and \emph{Dalphon}, and \emph{Aspatha}, and \emph{Poratha}, and \emph{Adalia}, and \emph{Aridatha}, and \emph{Parmashta}, and \emph{Arisai}, and \emph{Aridai}, and \emph{Vajezatha},
      & Och de dr{\"a}pte \emph{Kedak}, \emph{Ir-Fittim}, \emph{Aquila}, \emph{d{\"o}rrvaktarna}, \emph{Amarja}, \emph{Bered}, \emph{vidare Bet-Hadt}, \emph{Berota}, \emph{Gat-Rimmon},
      & Och de dr{\"a}pte \emph{Parsandata} \emph{Dalefon} och \emph{Aspata} \emph{Porata} \emph{Adalja} \emph{Aridata} \emph{Parmasta} \emph{Arisai} \emph{Aridai} \emph{Vajsata}
      & Och \emph{Parsandata}, \emph{Dalefon}, \emph{Aspata}, \emph{Porata}, \emph{Adalja}, \emph{Aridata}, \emph{Parmasta}, \emph{Arisai}, \emph{Aridai} och \emph{Vajsata},
      &\emph{Parshandatha}: 1, \emph{Dalphon}: 1, \emph{Aspatha}: 1, \emph{Poratha}: 1, \emph{Adalia}: 1, \emph{Aridatha}: 1, \emph{Parmashta}: 1, \emph{Arisai}: 1, \emph{Aridai}: 1, \emph{Vajezatha}: 1
    \\ \hline
  \end{tabular}
  \caption{Examples of order-preserving lexicon-aware translation for English to Swedish. The frequency of the named entities are the number of occurrences each named entity appears in the whole dataset; for example, all named entities in the last sentence only appear in the test set once, and do not appear in the training data.}
  \label{table:lexicon_results1}
\end{table*}
\subsection{Ablation Study on Target Training Data}
We use full training data from all rich-resource languages,
and we vary the amount of training data in Swedish, our low-resource language,
spanning from one tenth to full length uniformly.
We duplicate the subset to ensure all training sets,
though having a different number of unique
sentences, have the same number of total sentences.

\textbf{\ul{Power-law relationship is observed between the performance and the                                                                               
    amount of training data in low-resource language:}}
Figure \ref{fig:curve_multi} shows how BLEU scores
vary logarithmically with the number of
unique sentences in the low-resource
training data. It follows a linear pattern
for single-source single-target
translation from English to Swedish
as shown in Figure \ref{fig:curve_single}.
We also observe a linear pattern for the multi-source
multi-target case, though more uneven
in Figure \ref{fig:curve_multi}.
The linear pattern with BLEU scores against
the logarithmic data shows the power-law relationship between
the performance in translation and
the amount of low-resource training data.
Similar power-law relationships are also found in past research and contemporary literature \cite{turchi2008learning, hestness2017deep}.

\textbf{\ul{We achieve reasonably good BLEU scores
    using one fifth of random samples:}} For
the multi-source multi-target case, we find that using one fifth
of the low-resource training data gives reasonably good BLEU scores
as shown in Figure \ref{fig:curve_multi}.
This is helpful when we have
little low-resource data.
For translation into low-resource language,
the experts only need to translate a small amount of seed
data before passing it to our system \footnote{Note that
using nine tenth of random samples yields higher
performance than using full data, but it may not be generalized
to other datasets.}.

\subsection{Order-preserving Lexiconized NMT}
We devise a mechanism to build
a parallel lexicon table across twenty-three European
languages using very little data and zero manual work.
A few lexicon examples are shown in Table \ref{table:lexicon_examples}.
We first extract named entities from the English Bible
\cite{manning2014stanford} and combine them with English biblically named entities from multiple
sources \cite{easton1897eastons, nave1903nave, smith1967smith, hitchcock1874hitchcock, rice2015people}.
Secondly, we carefully automate the filtering process to obtain a clean English
lexicon list. Using this list as the seed, we build a parallel lexicon
table across all twenty-three languages through
fast-aligning \cite{dyer2013simple}. 
The final parallel lexicon table has 2916 named entities. In
the translation task into low-resource language,
we assume that the experts first translate these lexicon entries,
and then translate approximately one fifth random sentences
before we train our NMT. If necessary, the experts evaluate
and correct translations before releasing the final
translations to the low-resource language
community. We aim to reduce human effort in post-editing and increase
machine accuracy. After labeling named entities in each sentence pair in
order, we train and obtain good translation results.

\textbf{\ul{We observe 60.6\% accuracy in human evaluation where our translations are parallel to human translations:}} In Table \ref{table:lexicon_results1}, we show some examples of machine
translated text, we also show the expected correct translations for comparison. Not only the named entities are correctly mapped, but also
the ordering of the subject and the object is preserved.
In a subset of our test set, we conduct
human evaluation on 320 English-Swedish results to
rate the translations into three categories:
accurate (parallel to human translation), almost accurate 
(needing minor corrections) and inaccurate. More precisely, 
each sentence is evaluated using three criteria: correct set 
of named entities, correct positioning of named entities, and accurate 
meaning of overall translation. 
If a sentence achieves all three, then it is termed as accurate;  
if either a name entity is missing or its position is
 wrong, then it is termed as almost accurate (needing 
minor correction); if the meaning of the sentence 
is entirely wrong, then it is inaccurate. 
Our results are 60.6\% accurate, 33.8\% needing minor corrections, and 5.6\% inaccurate. Though human evaluation
carries bias and the sample is small,
it does give us perspective on the performance of our model.

\textbf{\ul{Order-preservation performs well especially when the named entities
    are rare words:}} In Table \ref{table:lexicon_results1}, NMT
without order-preservation lexiconized treatment
performs well when named entities
are common words, but fails to predict
the correct set of named entities and
their ordering when named entities are rare words.
The last column shows the
number of occurrences of each named entity. For the last example,
there are many named entities that only occur in data once,
which means that they never appear in training and
only appear in the test set. The normal NMT without
order-preservation lexiconized treatment
predicts the wrong set of
named entities with the wrong ordering. Our lexiconized
order-preserving NMT, on the contrary, performs
well at both the head and tail of the distribution,
predicts the right set of named entities with the right ordering.

\textbf{\ul{Prediction with longer sentences and many named entities
    are handled well:}} In Table \ref{table:lexicon_results1}, 
we see that normal NMT
without order-preservation lexiconized treatment 
performs well with short sentences
and few named entities in a sentence. 
But as the number of the name entities per sentence increases,
especially when the name entities are rare unknowns 
as discussed before, normal NMT cannot make correct
prediction of the right set of name entities 
with the correct ordering \ref{table:lexicon_results1}. 
Our lexiconized order-preserving NMT, 
on the contrary, gives very high accuracy when there are
many named entities in the sentence 
and maintains their correct ordering. 

\textbf{\ul{Trimming the lexicon list that keeps the tail helps to increase BLEU scores:}} Different from most of the previous lexiconized NMT works where BLEU scores never increase \cite{wang2017sogou},
our BLEU scores show minor improvements. BLEU score for German-Swedish translation increases from 35.8
to 36.6 in Table \ref{table:summary_lexicon}. 
As an attempt to increase our BLEU scores even further,
we conduct two more experiments. In one setting, we keep only the tail of the lexicon table
that occur in the Bible once. In another setting, we keep only a manual selection of lexicons.
Note that this is the only place where manual work is involved and is not essential. 
There are minor improvements in BLEU scores in both cases. 

\textbf{\ul{33.8\% of the translations require minor corrections: }}
The sentence length for these translations that require minor corrections 
is often longer. We notice that
some have repetitions that do not affect meaning, but need to be trimmed. Some
have the under-prediction problem where certain named entities in the source sentence
never appear; in this case, missing named entities need to be added.
Some have minor issues with plurality and tense. 
We show a few examples of the translations that need minor corrections
in the appendices for reference. Typically, sentences with longer sentence length
and more complicated named entity relationships require minor corrections to
achieve high translation quality. 

\section{Conclusion and Future Directions} \label{conclusion}
We present our order-preserving translation system for cross-lingual 
learning in European languages. We examine 
three issues that are important to translation into
low-resource language: the lack of low-resource data, effective
cross-lingual transfer, and the variable-binding problem.

Firstly, we add the source and the target family labels 
in training and examined intra-family and inter-family effects. 
We find that training on multiple families,
more specifically, training on two neighboring
families nearest to the low-resource language
improves BLEU scores to a reasonably good level.
Secondly, we devise a
rigorous ablation study and 
show that we only need a small portion of the
low-resource target data to 
produce reasonably good BLEU scores.
Thirdly, to address
the variable-binding problem, we build a 
parallel lexicon table across twenty-three European
languages and design a novel order-preserving
named entity translation method by tagging named entities 
in each sentence in order. We achieve
reasonably good 
quantitative and qualitative improvements
in a preliminary study.  

The order-preserving named entity translation labels
named entities in order. Since there are relatively less number of 
long sentences with many named entities than short sentences with
few named entities, underprediction of
named entities in long sentences may occur.
To seek solution to the underprediction
problem, we are looking at randomized
labeling of the named entities. Moreover, our
order-preserving named entity translation method works well
with a fixed pool of named entities in any static 
document known in advance. This is due to our unique use cases for
applications like translating water, sanitation and hygiene (WASH)
guidelines written in the introduction. We devise our method
to ensure high accuracy targeting translating named entities in static
document known in advance.
However, researchers may need to translate dynamic document to low-resource
language in real-time. We are actively researching 
into the dynamic timely named entity discovery with high accuracy.

We are actively extending our work to cover more world languages,
more diverse domains, and more varied sets of datasets to show our
methods are generalizable. Since our experiments shown in this paper
are using European languages, we are also interested
on non-European languages like Arabic, Indian, Chinese, Indonesian and many others to
show that our model is widely generalizable. We also expect to discover
interesting research ideas exploring a wider universe of linguistically
dissimilar languages. 

Our work is helpful for translation into low-resource language,
where human translators only need to translate a few lexicons and 
a partial set of data before passing it to
our system. Human translators may also be needed during post-editing
before a fully accurate translation is released. Our future goal is to 
minimize the human correction efforts and to present high quality
translation timely.

We would also like to work on real world low-resource tribal languages
where there is no or little training data. Translation using 
limited resources and data 
in these tribal groups that fits with the
culture-specific rules 
will be very important \cite{levin1998interlingua}. 
Real world low-resource
languages call for cultural-aware translation.

\section*{Acknowledgments}
We would like to thank Prof. Eduard Hovy for his insights on the topic and helpful suggestions. We would also like to thank Prof. Michael Cysouw for his generous sharing of the massive Bible corpus. 

\bibliography{emnlp2018}
\bibliographystyle{acl_natbib_nourl}

\clearpage
\begin{appendices}
  \label{sec:supplemental} 
  \section{Supplemental Materials}
  We show a few examples
  of full tables of NMT translation results in the next two pages. We also show some qualitative examples that need minor corrections for reference.  

  \begin{table*}[h!]
  \small
  \centering
  \begin{tabularx}{0.7\textwidth}{|p{\dimexpr.077\linewidth-2\tabcolsep-1.3333\arrayrulewidth}|X|X|X|X|X|X|X|X|X|X|X|} \hline
    lan & de & dn & dt & en & es & fr & it & no & po & ro & sw\\ \hline \hline
    de & N.A. & 43.6 & 49.2 & 51.1 & 46.4 & 47.2 & 43.4 & 47.3 & 47.1 & 44.1 & 42.2\\ \hline
    dn & 45.5 & N.A. & 43.7 & 45.1 & 43.7 & 44.4 & 41.3 & 46.2 & 42.9 & 42.0 & 41.8\\ \hline
    dt & 49.9 & 41.9 & N.A. & 51.2 & 45.3 & 46.1 & 42.4 & 44.2 & 46.5 & 43.1 & 41.6\\ \hline
    en & 48.5 & 42.0 & 48.5 & N.A. & 47.0 & 46.2 & 43.3 & 43.8 & 46.3 & 42.3 & 40.2\\ \hline
    es & 46.8 & 41.8 & 46.0 & 48.4 & N.A. & 46.2 & 42.5 & 44.1 & 47.1 & 43.7 & 41.0\\ \hline
    fr & 44.9 & 41.1 & 42.9 & 45.4 & 45.1 & N.A. & 41.1 & 43.6 & 45.1 & 48.5 & 39.8\\ \hline
    it & 46.7 & 40.6 & 45.9 & 47.3 & 44.8 & 46.5 & N.A. & 43.5 & 46.3 & 42.5 & 39.8\\ \hline
    no & 47.8 & 46.3 & 45.4 & 47.7 & 45.1 & 46.6 & 41.9 & N.A. & 44.5 & 43.5 & 44.0\\ \hline
    po & 48.3 & 42.4 & 45.9 & 49.2 & 47.2 & 46.5 & 43.0 & 44.5 & N.A. & 44.5 & 40.6\\ \hline
    ro & 43.7 & 40.4 & 43.5 & 44.8 & 43.7 & 50.7 & 41.1 & 42.0 & 44.3 & N.A. & 38.5\\ \hline
    sw & 45.8 & 43.0 & 44.0 & 45.9 & 44.2 & 44.9 & 40.8 & 46.6 & 44.3 & 42.6 & N.A.\\ \hline
  \end{tabularx}
   \caption{Germanic and romance families multi-source multi-target translation.}
   \label{table:families_germanic_romance}
\end{table*}
\begin{table*}[h!]
  \small
  \centering
  \begin{tabularx}{\textwidth}{|X|X|X|X|X|X|X|X|X|X|X|X|X|X|X|X|X|} \hline
    lan & cr & cz & de & dn & dt & en & es & fr & it & no & po & po & ro & ru & sw & uk\\ \hline \hline
    cr & N.A. & 40.9 & 44.5 & 42.3 & 43.9 & 46.8 & 44.9 & 45.0 & 41.4 & 44.1 & 41.9 & 43.9 & 41.9 & 43.6 & 40.4 & 41.0\\ \hline
    cz & 36.4 & N.A. & 48.7 & 44.3 & 47.7 & 50.8 & 46.7 & 46.0 & 44.2 & 46.9 & 48.5 & 47.1 & 45.2 & 46.6 & 42.2 & 44.4\\ \hline
    de & 36.0 & 44.0 & N.A. & 45.0 & 49.4 & 52.2 & 48.0 & 46.9 & 45.5 & 48.6 & 45.7 & 48.3 & 45.3 & 47.9 & 42.5 & 44.7\\ \hline
    dn & 35.0 & 41.2 & 47.1 & N.A. & 45.4 & 48.4 & 45.8 & 45.6 & 42.0 & 47.3 & 42.8 & 44.4 & 42.9 & 45.1 & 42.6 & 42.0\\ \hline
    dt & 35.3 & 43.4 & 51.0 & 45.5 & N.A. & 52.3 & 46.2 & 46.0 & 45.4 & 47.3 & 45.9 & 47.5 & 44.9 & 46.9 & 42.8 & 44.0\\ \hline
    en & 36.4 & 43.2 & 50.3 & 44.6 & 48.9 & N.A. & 47.7 & 47.0 & 45.3 & 47.0 & 46.3 & 48.0 & 43.8 & 47.2 & 41.8 & 44.6\\ \hline
    es & 36.7 & 43.5 & 48.8 & 44.7 & 47.6 & 50.4 & N.A. & 47.3 & 44.9 & 47.0 & 44.3 & 49.4 & 45.6 & 46.5 & 42.4 & 44.0\\ \hline
    fr & 35.8 & 42.7 & 47.6 & 43.6 & 45.3 & 49.2 & 45.9 & N.A. & 44.0 & 45.2 & 43.1 & 46.8 & 48.0 & 46.0 & 41.6 & 42.6\\ \hline
    it & 35.5 & 44.0 & 48.4 & 43.8 & 47.2 & 50.4 & 45.5 & 47.0 & N.A. & 45.6 & 45.3 & 47.4 & 46.3 & 46.7 & 41.6 & 43.1\\ \hline
    no & 36.2 & 44.0 & 49.3 & 46.1 & 47.5 & 49.9 & 47.2 & 47.6 & 44.0 & N.A. & 45.5 & 46.1 & 45.2 & 47.1 & 44.5 & 43.2\\ \hline
    po & 35.7 & 47.1 & 48.7 & 44.5 & 48.9 & 51.0 & 46.5 & 47.2 & 44.8 & 46.4 & N.A. & 48.0 & 44.4 & 47.6 & 41.6 & 45.0\\ \hline
    po & 37.2 & 44.1 & 49.1 & 45.1 & 47.9 & 50.1 & 48.4 & 46.8 & 44.9 & 46.8 & 45.8 & N.A. & 45.4 & 47.2 & 43.1 & 43.9\\ \hline
    ro & 35.4 & 42.6 & 46.0 & 42.7 & 45.0 & 48.1 & 45.6 & 49.4 & 43.0 & 45.4 & 42.8 & 46.2 & N.A. & 44.4 & 42.3 & 41.8\\ \hline
    ru & 36.4 & 44.7 & 48.5 & 44.8 & 47.5 & 49.9 & 47.0 & 46.4 & 45.1 & 47.0 & 46.2 & 46.9 & 45.6 & N.A. & 42.2 & 44.4\\ \hline
    sw & 35.5 & 42.6 & 46.8 & 44.5 & 45.1 & 47.6 & 45.3 & 45.9 & 42.1 & 47.2 & 43.7 & 45.8 & 44.1 & 45.8 & N.A. & 42.8\\ \hline
    uk & 34.5 & 44.2 & 48.7 & 43.0 & 47.5 & 50.4 & 46.2 & 45.3 & 44.7 & 46.6 & 46.0 & 46.0 & 43.8 & 47.8 & 42.1 & N.A.\\ \hline
  \end{tabularx}
  \caption{Germanic, romance and slavic families multi-source multi-target translation.}
  \label{table:families3}
\end{table*}
\begin{table*}[h!]
  \small
  \centering
  \hspace*{-1cm}
  \begin{tabularx}{1.12\textwidth}{|X|X|X|X|X|X|X|X|X|X|X|X|X|X|X|X|X|X|X|X|X|X|X|} \hline
    lan & ab & cr & cz & de & dn & dt & en & es & fn & fr & gk & hg & it & ln & no & po & po & ro & ru & sw & uk & ws\\ \hline \hline
    ab & N.A. & 35.2 & 42.1 & 46.7 & 43.2 & 46.9 & 48.8 & 46.5 & 41.3 & 47.3 & 50.9 & 41.0 & 44.0 & 37.4 & 44.0 & 44.8 & 46.4 & 44.4 & 46.3 & 41.3 & 43.7 & 45.4\\ \hline
    cr & 41.1 & N.A. & 41.6 & 47.5 & 43.4 & 44.6 & 46.5 & 45.1 & 40.4 & 44.9 & 49.6 & 39.8 & 41.9 & 37.6 & 44.7 & 44.3 & 44.9 & 44.1 & 45.7 & 41.5 & 43.6 & 44.3\\ \hline
    cz & 43.9 & 35.7 & N.A. & 49.3 & 44.2 & 48.5 & 49.7 & 46.8 & 43.9 & 47.1 & 53.6 & 42.8 & 44.0 & 41.4 & 46.1 & 48.8 & 46.7 & 45.0 & 49.6 & 42.9 & 45.8 & 46.6\\ \hline
    de & 43.2 & 36.7 & 44.8 & N.A. & 46.9 & 49.7 & 52.4 & 49.3 & 44.1 & 47.9 & 53.8 & 42.7 & 45.1 & 39.9 & 47.4 & 46.0 & 47.6 & 45.2 & 48.6 & 44.3 & 46.6 & 48.4\\ \hline
    dn & 42.1 & 35.9 & 43.1 & 48.1 & N.A. & 46.4 & 49.4 & 46.6 & 42.0 & 45.5 & 51.6 & 41.0 & 42.4 & 38.6 & 47.6 & 43.9 & 46.8 & 44.8 & 45.7 & 42.8 & 43.0 & 45.0\\ \hline
    dt & 42.8 & 35.2 & 44.9 & 50.7 & 44.4 & N.A. & 52.5 & 47.4 & 43.5 & 47.3 & 54.9 & 43.2 & 45.5 & 40.3 & 46.4 & 46.3 & 49.1 & 45.4 & 49.1 & 42.3 & 45.8 & 47.5\\ \hline
    en & 42.9 & 34.9 & 44.5 & 49.6 & 45.7 & 49.8 & N.A. & 47.4 & 43.2 & 47.6 & 54.2 & 42.9 & 44.5 & 40.3 & 47.0 & 46.0 & 48.4 & 45.1 & 48.4 & 43.9 & 47.9 & 48.5\\ \hline
    es & 43.2 & 36.1 & 44.1 & 48.6 & 43.3 & 47.8 & 49.5 & N.A. & 42.0 & 47.3 & 53.6 & 42.8 & 45.2 & 39.3 & 46.4 & 45.2 & 49.2 & 46.1 & 48.2 & 43.1 & 45.2 & 46.4\\ \hline
    fn & 41.3 & 33.9 & 42.8 & 47.8 & 43.5 & 46.6 & 48.7 & 44.9 & N.A. & 44.6 & 51.2 & 41.3 & 42.3 & 38.3 & 43.7 & 44.2 & 45.2 & 43.0 & 46.1 & 40.8 & 43.7 & 44.4\\ \hline
    fr & 42.7 & 33.9 & 41.5 & 49.0 & 43.6 & 46.0 & 48.5 & 46.8 & 42.1 & N.A. & 52.6 & 41.1 & 44.2 & 38.8 & 45.4 & 43.7 & 46.4 & 48.8 & 47.6 & 42.1 & 43.9 & 45.5\\ \hline
   gk & 43.3 & 35.7 & 43.9 & 50.0 & 46.3 & 49.0 & 51.4 & 48.0 & 43.8 & 47.4 & N.A. & 43.1 & 43.6 & 40.4 & 45.3 & 46.7 & 47.7 & 44.6 & 48.4 & 41.7 & 45.5 & 47.8\\ \hline
    hg & 41.8 & 34.9 & 42.7 & 47.2 & 43.8 & 46.7 & 48.9 & 45.9 & 42.7 & 46.0 & 52.4 & N.A. & 43.5 & 38.9 & 44.8 & 45.0 & 45.7 & 44.0 & 47.6 & 42.0 & 44.5 & 45.5\\ \hline
    it & 43.0 & 34.7 & 43.5 & 48.7 & 43.7 & 47.8 & 50.6 & 46.3 & 42.6 & 46.8 & 52.7 & 41.4 & N.A. & 38.8 & 45.4 & 44.4 & 46.9 & 44.9 & 47.7 & 41.8 & 44.8 & 46.5\\ \hline
    ln & 40.4 & 33.6 & 41.4 & 46.0 & 42.8 & 45.4 & 48.4 & 44.6 & 41.3 & 44.5 & 51.4 & 40.8 & 42.5 & N.A. & 43.7 & 44.7 & 45.5 & 42.9 & 46.0 & 39.3 & 42.3 & 43.8\\ \hline
    no & 43.0 & 35.6 & 44.6 & 48.3 & 47.1 & 47.3 & 49.8 & 47.1 & 44.1 & 46.9 & 52.2 & 42.9 & 43.7 & 39.0 & N.A. & 46.0 & 46.6 & 45.3 & 48.8 & 44.6 & 46.1 & 46.9\\ \hline
    po & 43.5 & 35.6 & 47.4 & 50.3 & 44.6 & 48.0 & 50.8 & 46.8 & 44.4 & 46.8 & 53.9 & 43.1 & 44.1 & 40.9 & 46.4 & N.A. & 47.5 & 44.7 & 49.3 & 43.0 & 46.8 & 47.3\\ \hline
    po & 44.2 & 36.1 & 44.1 & 48.9 & 45.1 & 47.7 & 50.8 & 48.2 & 44.9 & 47.9 & 53.6 & 42.9 & 44.5 & 39.6 & 46.0 & 46.4 & N.A. & 45.6 & 48.5 & 41.8 & 46.4 & 46.8\\ \hline
    ro & 44.0 & 34.8 & 41.7 & 46.9 & 43.1 & 45.8 & 48.5 & 46.1 & 42.0 & 49.2 & 50.6 & 39.9 & 43.3 & 37.1 & 45.5 & 44.0 & 45.8 & N.A. & 46.5 & 41.6 & 43.4 & 44.9\\ \hline
    ru & 43.7 & 36.9 & 45.6 & 50.1 & 44.7 & 48.2 & 50.2 & 47.2 & 43.3 & 46.9 & 53.4 & 42.5 & 45.2 & 40.0 & 47.5 & 45.8 & 46.6 & 46.3 & N.A. & 42.6 & 47.0 & 47.1\\ \hline
    sw & 42.4 & 34.5 & 42.5 & 48.0 & 45.5 & 45.9 & 47.9 & 45.8 & 43.0 & 45.3 & 51.4 & 41.8 & 43.4 & 37.3 & 46.8 & 44.5 & 45.3 & 43.8 & 46.5 & N.A. & 43.9 & 45.9\\ \hline
    uk & 42.8 & 35.8 & 45.0 & 49.9 & 44.3 & 48.5 & 50.5 & 45.9 & 43.5 & 47.0 & 53.8 & 43.1 & 44.0 & 40.0 & 46.0 & 47.5 & 47.5 & 45.3 & 50.1 & 43.3 & N.A. & 47.2\\ \hline
    ws & 42.8 & 36.3 & 44.5 & 49.4 & 43.1 & 48.3 & 51.5 & 46.2 & 43.2 & 46.8 & 53.0 & 43.1 & 44.7 & 39.1 & 45.6 & 45.6 & 47.4 & 45.0 & 49.0 & 41.6 & 45.5 & N.A.\\ \hline
  \end{tabularx}
  \caption{Eight European families family-labelled multi-source multi-target translation.}
  \label{table:bigeuropean_family_labelled}
\end{table*}    

\begin{table*}[]
  \small
  \centering
  \begin{tabular}{ |p{3.1cm}|p{3.1cm}|p{3.1cm}|p{3.1cm}|p{1.8cm}| } \hline
    Source Sentence
    & NMT Translation without Order Preservation (Before)
    & NMT Translation with Order Preservation (After)
    & Correct Target Translation
    & Frequency of Named Entities
    \\ \hline \hline
    And \emph{David} fled, and escaped, and came to \emph{Samuel} at \emph{Ramah}, and told him all that \emph{Saul} had done to him .
    & Och \emph{David} flydde, och han flydde, och n{\"a}r han hade kommit till \emph{Rama}, ber{\"a}ttade han f{\"o}r honom allt vad \emph{ Saul} hade gjort.
    & D{\aa} flydde \emph{David} bort och kom till \emph{Samuel} och kom till \emph{Samuel} i \emph{Rama} och ber{\"a}ttade f{\"o}r honom al lt vad \emph{Saul} hade gjort mot honom ;
    & N{\"a}r \emph{David} nu hade flytt och kommit undan, begav han sig till \emph{Samuel} i \emph{Rama} och omtalade f{\"o}r denne allt va d \emph{Saul} hade gjort honom.
    & \emph{David}: 1134, \emph{Samuel}: 140, \emph{Ramah}: 39, \emph{Saul}: 424
    \\ \hline
    And \emph{Joseph} made it a law until this day on the land of \emph{Egypt}, the fifth part is for \emph{Pharaoh}; only the land of the p riests being excepted; it did not become \emph{Pharaoh}'s.
    & Och \emph{Josef} gjorde ett lag {\"a}nda till denna dag, i \emph{Egyptens} land, den femte allenast f{\"o}r \emph{Farao}; allenast int ill den andra av pr{\"a}sternas land gjorde man icke n{\aa}got som f{\"o}rtj{\"a}nade \emph{Farao}.
    & Och \emph{Josef} gjorde en lag {\"a}nda till denna dag i \emph{Egyptens} land, den femte lotten som utgjorde ett stycke {\aa}t \emph{F arao} Allenast prestennas land utgjorde ej av \emph{Farao}.
    & S{\aa} gjorde \emph{Josef} det till en stadga, som {\"a}nnu i dag g{\"a}ller f{\"o}r \emph{Egyptens} jord, att man skulle giva femtede len {\aa}t \emph{Farao}. Allenast pr{\"a}sternas jord blev icke \emph{Faraos} egendom.
    &\emph{Joseph}: 251, \emph{Egypt}: 650, \emph{Pharaoh}: 272
    \\ \hline
    And his sons carried him to the land of \emph{Canaan}, and buried him in the cave of the field of \emph{Machpelah}; the field which \emph{Abraham} bought for a burying place from \emph{Ephron} the \emph{Hittite}, before \emph{Mamre}.
    & Och hans s{\"o}ner f{\"o}rde honom in i \emph{Kanaans} land och buro honom p{\aa} marken p{\aa} marken. Den {\aa}ker som \emph{Abraham } hade k{\"o}pt till plundring fr{\aa}n R{\"o}da havet , f{\"o}r \emph{Mamre}.
    & Och hans s{\"o}ner f{\"o}rde honom till \emph{Kanaans} land , och man begrov honom i \emph{Makpela} land ; den{\aa}ker som \emph{Abrah am} hade k{\"o}pt fr{\aa}n \emph{hetiten} \emph{Efron} till begravningsplats .
    & hans s{\"o}ner f{\"o}rde honom till \emph{Kanaans} land och begrovo honom i grottan p{\aa} {\aa}kern i \emph{Makpela}, den {\aa}ker so m \emph{Abraham} hade k{\"o}pt till egen grav av \emph{hetiten} \emph{Efron}, gent emot \emph{Mamre}.
    &\emph{Canaan}: 94, \emph{Machpelah}: 5, \emph{Abraham}: 248, \emph{Ephron}: 13, \emph{Hittite}: 34, \emph{Mamre}: 10
    \\ \hline
  \end{tabular}
  \caption{Examples of order-preserving lexicon-aware translation that needs minor corrections for English to Swedish. The frequency of the  named entities are the number of occurrences each named entity appears in the whole dataset. }
  \label{table:lexicon_results2}
\end{table*}

\end{appendices}

\end{document}